# A Novel Framework For Text Detection From Natural Scene Images With Complex Background.

Basavaraj Kaladagi , Dr. Jagadeesh Pujari
basu888@gmail.com, jaggudp@hotmail.com
SDM College of Engineering and Tech, Dharwad (VTU).

**Abstract**

*Recognizing texts from camera images is a known hard problem because of the difficulties in text detection from the varied and complicated background. In this paper we propose a novel and efficient method to detect text region from images with complex background using Wavelet Transforms. The framework uses Wavelet Transformation of the original image in its grayscale form followed by Sub-band filtering. Then Region clustering technique is applied using centroids of the regions, further Bounding box is fitted to each region thus identifying the text regions.*

*This method is much sophisticated and efficient than the previous methods as it doesn't stick to a particular font size of the text thus, making it generalized. The sample set used for experimental purpose consists of 50 images with varying backgrounds. Images with edge prominence are considered. Furthermore, our method can be easily customized for applications with different scopes.*

## 1. Introduction

In recent years, we have created many still-images and videos using digital cameras, digital camcorders and cellular-phone cameras. Texts in these images contain very important information about locations and road signs. If we could recognize these texts accurately in real time, we can design artificial vision systems for assisting auto-navigation of vehicles and vision impaired, video indexing and retrieval systems, text translation systems, and spam-mail filtering systems, and so on.
The research on Text Detection from Natural Scene Images, towards a System for Visually Impaired Persons [1] proposing a system that reads the text encountered in natural scenes with the aim to provide assistance to the visually impaired persons. Methods are proposed which present a novel image operator that seeks to find the value of stroke width for each image pixel, and demonstrate its use on the task of text detection in natural images [2]. Detection of text in color images of complex colored background is done through an efficient automatic text detection method-fusing multi-feature [3]. A low complexity method for detection of text regions in natural images which is designed for mobile applications (e.g. unmanned or hand-held devices) in which computational and energy resources are limited, it focuses on the detection of text on signs and it relies on the properties of the background of the text region, as opposed to the text itself [4]. Text detection from natural scene images can also be done based on the intensity information of the images. This method is composed of gray value stretching and binarization by an average intensity of the image. It is appropriate to extract texts from complex backgrounds [5]. A new method for text detection and recognition in natural scene images is presented in which detection process, color, texture, and OCR statistic features are combined in a coarse-to-fine framework to discriminate texts from non-text patterns. In this approach, color feature is used to group text pixels into candidate text lines. Texture feature is used to capture the ''dense intensity variance'' property of text pattern. Statistic features from OCR (Optical Character Reader) results are employed to further reduce detection false alarms empirically. After the detection process, a restoration process is used [6]. In one of the methods text detection is done using multiscale texture segmentation and spatial cohesion constraints, then cleaned up and extracted using a histogram-based binarisation algorithm [7]. A frame work is proposed to detect text using text intensity and shape features, this framework uses the Niblack algorithm to threshold images and group components into regions with commonly used geometry features. The intensity filter considers the overlap between the intensity histogram of a component and that of its adjoining area. For non-text regions, it is found that this overlap is large, and hence can prune out components with large values of this measure. The shape filter, on the other hand, deletes regions whose constituent components come from a same object, as most words consist of different characters [8].

## 2. Proposed Method

In this section, we present a method to extract texts in natural scene images using Haar discrete wavelet transform (Haar DWT). The edge detection is accomplished by using 2-D Haar DWT and some of the non-text edges are removed using thresholding. Although the color component may differ in a text region, the information about color does not help extracting texts from images. If the input is a gray-level image, the image is processed directly starting at discrete wavelet transform. If the input is RGB image, it is converted to grayscale image and processed.

The discrete wavelet transform is a very useful tool for signal analysis and image processing, especially in multi-resolution representation. Two-dimensional discrete wavelet transform (2-D DWT) decomposes an input image into four sub-bands, one average component (LL) and three detail components (LH, HL, HH) as shown in Figure 2.1

| LL | HL |
|----|----|
| LH | HH |

Figure 2.1

The traditional edge detection filters can provide a similar result as well. However, 2-D DWT can detect three kinds of edges at a time while traditional edge detection filters cannot. The traditional edge detection filters detect three kinds of edges by using four kinds of mask operators. Therefore, processing time of the traditional edge detection filters is slower than 2-D DWT.

Three kinds of edges present in the detail component sub-bands but look unobvious (very small coefficients). DWT filters with Haar DWT, the detected edges become more obvious and the processing time decreases. The operation for Haar DWT is simpler than that of any other wavelets. It has been applied to image processing especially in multi-resolution representation.
By applying Haar DWT to input image, we can obtain various features about the original image as follows:
1. Average components are detected by the LL sub-band;
2. Vertical edges are detected by the HL sub-band;
3. Horizontal edges are detected by the LH sub-band;
4. Diagonal edges are detected by the HH sub-band.

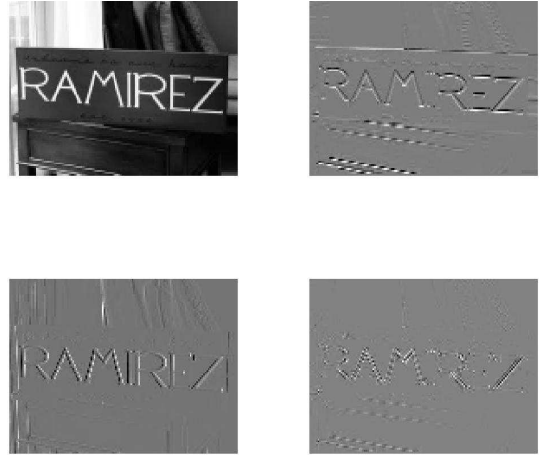

Figure 2.2

The figure 2.2 shows the horizontal, vertical and diagonal sub-band images obtained after applying dwt.

## 2.1 Proposed Method for Text Detection.

In this subsection, we use morphological operators and the logical AND operator to further removes the non-text regions. In text regions, vertical edges, horizontal edges and diagonal edges are mingled together while they are distributed separately in non-text regions. Since text regions are composed of vertical edges, horizontal edges and diagonal edges, we can determine the text regions to be the regions where those three kinds of edges are intermixed. Text edges are generally short and connected with each other in different orientation. Thus we apply logical AND operator to horizontal, vertical and diagonal sub-band image set to get the region of interest (ROI). The figure 2.1.1 shown below shows horizontal, vertical and diagonal pixels and the text region after applying AND operation to the sub-band images.

While applying haar wavelet transform, sigma value decides the amount of filtering done i.e. as the value increases the amount of non-text but textured region present in the image decreases thus pruning out the non-text region.

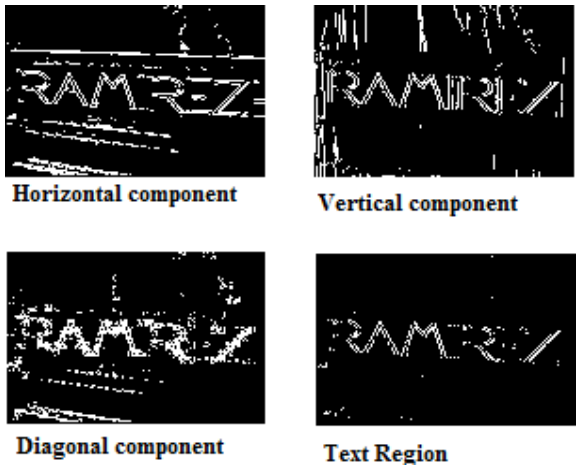

Figure 2.1.1

### 2.1.1 A Proposed method for Filtering

Even after this much of filtering there is a probability of getting non-text region which have textured features appearing in the image. To filter such background pixels a new method is proposed. Here we scan through the image and record the rows containing white pixel concentration above a particular threshold. These pixels form the part of text region after filtering.

### 2.2 Clustering and Region Growing

The co-ordinates of the pixels obtained after region filtering are subjected to clustering. Here we use subtractive clustering to extract cluster centers.

The subtractive clustering method assumes each data point is a potential cluster center and calculates a measure of the likelihood that each data point would define the cluster center, based on the density of surrounding data points. The algorithm does the following:
- Selects the data point with the highest potential to be the first cluster center
- Removes all data points in the vicinity of the first cluster center in order to determine the next data cluster and its center location
- Iterates on this process until all of the data is within the vicinity of a cluster center
- The subtractive clustering method is an extension of the mountain clustering method proposed by R. Yager.

### 2.2.1 Clustering Regions.

Once region are divided into clusters, the process of growing region is done using the cluster center of each region, the algorithm is as follows:

```
REGION_GROWING(Cluster_Centroids, Number_Of_Clusters)
    For every c € Cluster_Centroids
        Put a 2 * 2  Rectangle around c
        New_value=Calculate number of
                  pixels belonging to text
        percentIncrease=(NewValue-OldValue)*100/ NewValue;
        if  percentIncrease< 5
            Draw a BoundingBox across
            that region;
            break;
        else
            OldValue=NewValue;
        End if
    End for
Return;
```

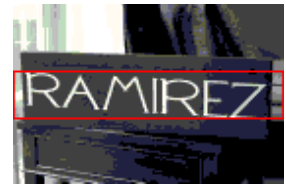

Figure 2.2.1.1

### 2.3 Thresholding

Once the text region is extracted from the image using the bounding box fitted to the text region, each region is subjected to thresholding, thresholding is done using Ostu's thresholding method which chooses the threshold to minimize the intraclass variance of the black and white pixels.

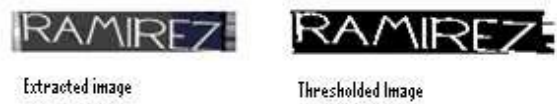

Figure 2.3.1

### 3. Character Separation

Once the text region is threshold , the characters have to be separated before sending them to OCR for recognition, as there is a little gap between every character on the text region, characters are separated using the simple and efficient Connected Component algorithm. Components are considered to be 4-connected. Connected Component is implemented so as to find the connectivity between white pixels. This leads to a problem, if the image supplied has black text with white background, to overcome this a strategy is developed to check the background of image which is as follows:

1. Put a 3 *3 rectangle at the corner of the image.
2. If the background is white and the characters are in black then we have to change the background color to black. Now we can label the objects by applyingsimple connected-component algorithm.
3. If the background is white, then complement of image is taken which is then subjected to connected- component algorithm.
4. Now each connected-component represents a character which is then supplied to OCR for recognition.

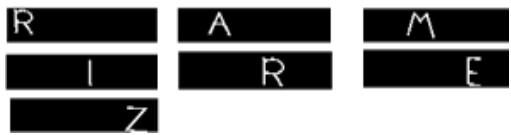

Figure 3.1

## 4. Conclusion

In this paper we have presented a novel framework to detect text region from natural scene images with complex background with a fixed font style.
The method uses discrete wavelet transform to get the sub-bands and region of interest detection is done using a Proposed method, which is based on the concentration of texture features, got by applying morphological AND operator. This method has a good efficiency for images containing objects with less texture properties.